\documentclass[letterpaper, 10 pt, conference]{ieeeconf}  

\IEEEoverridecommandlockouts    
\title{\LARGE \bf
FlatFusion: Delving into Details of Sparse Transformer-based Camera-LiDAR Fusion for Autonomous Driving}

\usepackage[T1]{fontenc} 
\usepackage{booktabs} 
\usepackage{graphicx}
\usepackage[breaklinks=true,bookmarks=true,colorlinks,pagebackref=true]{hyperref}
\usepackage{subcaption}
\usepackage{amsmath}
\usepackage{url}
\usepackage{adjustbox}
\usepackage{multirow}

\begin{document}

\author{Yutao Zhu$^{*,1}$, Xiaosong Jia$^{*,1}$, Xinyu Yang$^{2}$ and Junchi Yan$^{\dagger,1}$
\thanks{$^*$ Equal contributions. $^\dagger$ Correspondence author. The SJTU authors were in part supported by by NSFC (62222607) and Shanghai Municipal Science and Technology Major Project under Grant 2021SHZDZX0102.}%
\thanks{
$^{1}$ Y. Zhu, X. Jia, and J. Yan are with School of Computer Science and School of Artificial Intelligence, Shanghai Jiao Tong University, Shanghai, 200240, China. {\tt\small\{yt.zhu, jiaxiaosong, yanjunchi\}@sjtu.edu.cn}
}%
\thanks{$^{2}$ X. Yang is with Carnegie Mellon University, Pittsburgh, PA 15213, USA. {\tt\small xinyuya2@andrew.cmu.edu}}}

\maketitle
\thispagestyle{empty}
\pagestyle{empty}
\begin{abstract}

The integration of data from various sensor modalities (e.g. camera and LiDAR) constitutes a prevalent methodology within the ambit of autonomous driving scenarios.
Recent advancements in efficient point cloud transformers have underscored the efficacy of integrating information in sparse formats. When it comes to fusion, since image patches are dense in pixel space with ambiguous depth, it necessitates additional design considerations for effective fusion. In this paper, we conduct a comprehensive exploration of design choices for transformer-based sparse camera-LiDAR fusion. This investigation encompasses strategies for image-to-3D and LiDAR-to-2D mapping, attention neighbor grouping, single modal tokenizer, and  micro-structure of Transformer. By amalgamating the most effective principles uncovered through our investigation, we introduce FlatFusion, a carefully designed framework for sparse camera-LiDAR fusion. Notably, FlatFusion significantly outperforms state-of-the-art sparse Transformer-based methods, including UniTR, CMT, and SparseFusion, achieving 73.7 NDS on the nuScenes validation set with 10.1 FPS with PyTorch. 

\end{abstract}

\section{INTRODUCTION}
The pursuit of autonomous driving~\cite{hu2023uniad,yang2023llm4drive,li2023delving,yang2022learning,jia2024bench2drive,you2024bench2driver,yang2025raw2drive,yang2025resim,yang2025drivemoe} has necessitated the development of sophisticated 3D object detection systems~\cite{wang2021fcos3d,gao2020using} capable of accurately perceiving and interpreting the environment~\cite{ide-net,jia2022multi,zhangesdmotion}. At the forefront of this endeavor is the integration of multi-modal sensory inputs~\cite{liu2023bevfusion}, with a particular emphasis on the fusion of camera and LiDAR data. The synergy between these sensors, where cameras provide a wealth of semantic information and LiDARs offer precise geometric details, is a compelling avenue to improve the robustness and accuracy of autonomous driving systems~\cite{jia2022multi,jia2023hdgt,jia2024amp,wu2301policy,jia2021ide,jia2025drivetransformer,fan2025interleave,yang2025trajectory}. Despite the potential, the fusion of camera and LiDAR data presents a formidable challenge due to their distinct characteristics. Cameras capture dense, yet depth-ambiguous~\cite{li2023bevdepth}, 2D images, while LiDAR generates 3D point clouds that, although accurate, are sparse and less semantically rich~\cite{zhou2018voxelnet,zhang2025pointobb}. This discrepancy demands innovative approaches to integrate their complementary strengths.

Recent progress in the domain of sparse Transformers~\cite{sun2022swformer,liu2023flatformer,wang2023dsvt} has opened new avenues for efficient data integration in a format that is compatible with the sparsity of LiDAR data, demonstrating better effectiveness and scalability than SparseConv~\cite{spconv2022}. However, the dense nature of camera imagery requires careful consideration to enable a seamless fusion with the sparse LiDAR representation. Pioneering works~\cite{wang2023unitr,xie2023sparsefusion,yan2023cross} propose the general framework of spare fusion, leaving the specific design of each component unexplored. To fulfill this gap, \textbf{we delve into the design principles and architectural choices critical for the effective fusion of camera and LiDAR data through the lens of sparse Transformers via in-depth analysis, visualization, and ablations}. 

    \begin{figure}[!t]
        \centering
        \includegraphics[width=\columnwidth]{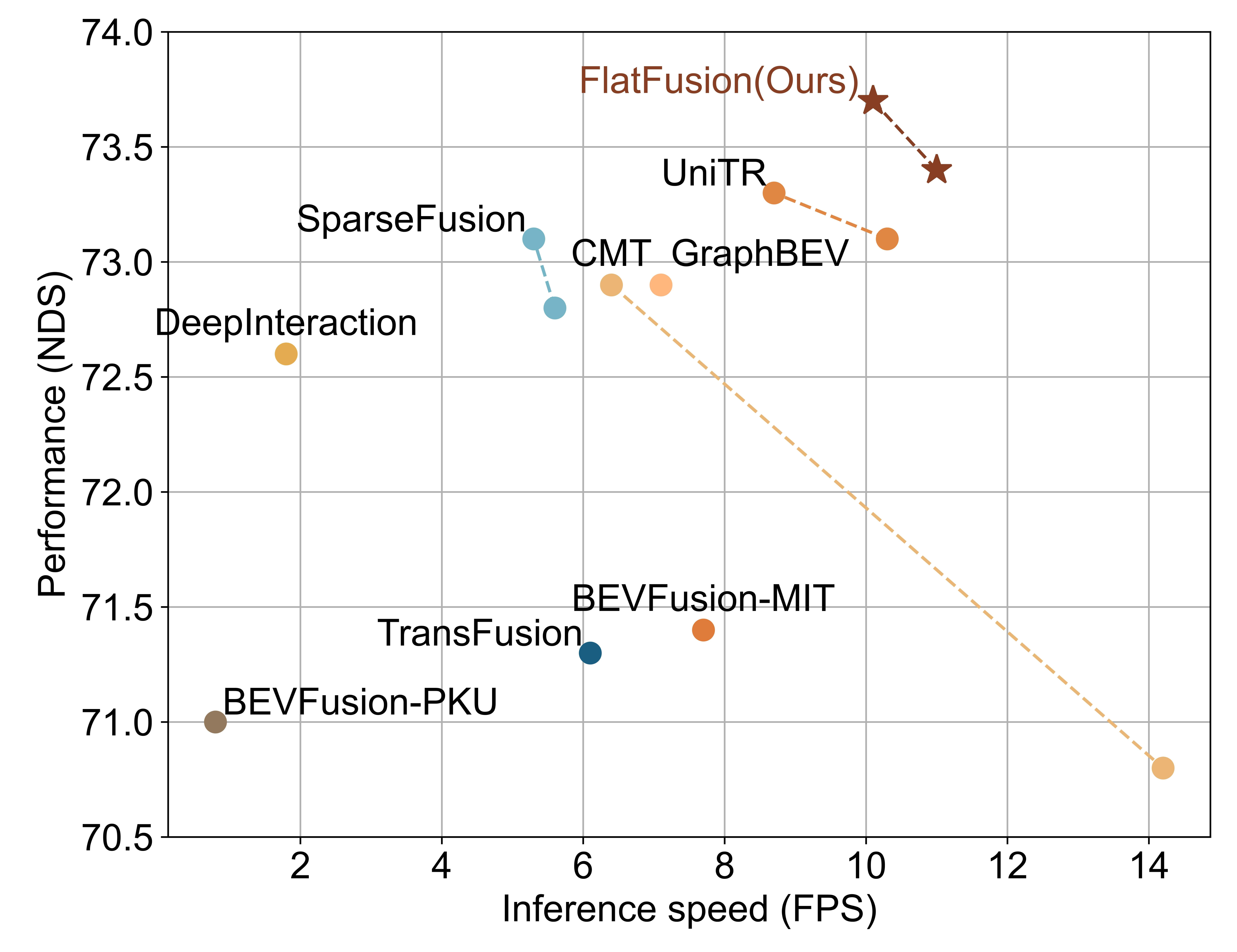}
        \caption{Comparison of performance on nuScenes \textit{validation} set.}
        \label{fig:ndsvsfps}
        \vspace{-2mm}
    \end{figure}

    \begin{figure*}[!t]
        \centering
        \includegraphics[width=\textwidth]{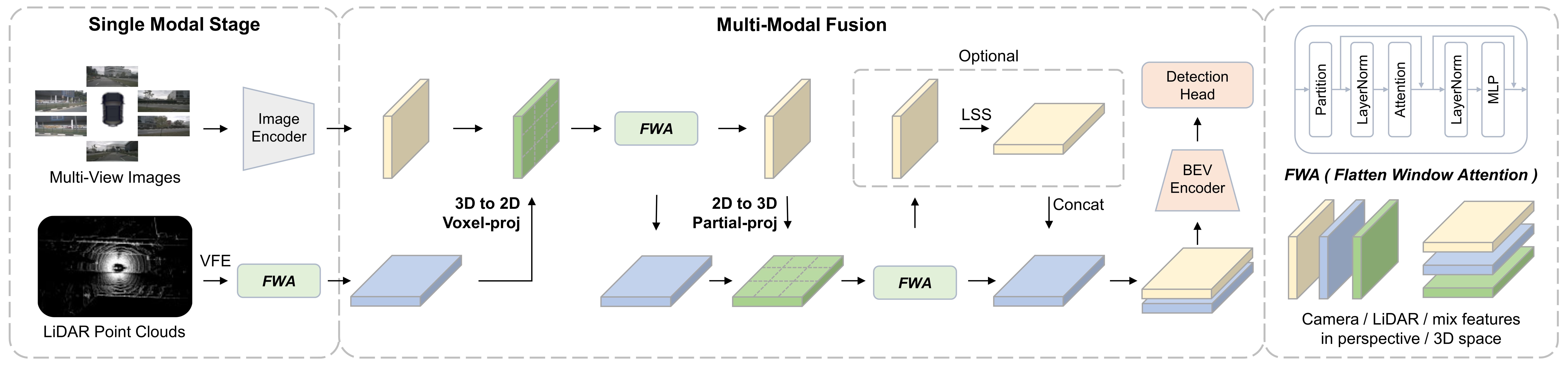}
        \caption{\textbf{General framework of sparse transformer-based camera-LiDAR fusion.} The input images and point cloud are fisrt tokenized by their modality-specific backbones. Then, tokens from both modalities are fused in different representation space (3D or pixel space) by projecting tokens from one modality to another. The fusion process is conducted by sparse window attention. Finally, the fused feature are fed into detection head.\label{fig:framework}}
    \end{figure*}
Our exploration encompasses most of ingredients for camera-LiDAR fusion, including image-to-3D and LiDAR-to-2D mapping techniques, attention neighbor grouping, single modal tokenizer and the micro-structure of Transformer. Through this investigation, we unveil design principles that are most conducive to sparse camera-LiDAR fusion, including light-weight image backbones and LiDAR backbones, Voxel-based 3D-to-2D fusion, partial projection for 2D-to-3D fusion, 3D-to-2D first order, Flatten Window partition algorithm, and PreNorm Transformer structure with 3D PE. Capitalizing on these insights, we introduce \textbf{FlatFusion}, a method carefully designed for sparse camera-LiDAR fusion. FlatFusion significantly outperforms existing state-of-the-art methods, including UniTR~\cite{wang2023unitr}, CMT~\cite{yan2023cross}, and SparseFusion~\cite{xie2023sparsefusion}, in terms of both accuracy and efficiency, as in Fig.~\ref{fig:ndsvsfps}. Notably, FlatFusion achieves a remarkable 73.7 NDS on the nuScenes validation set, while maintaining a processing speed of 10.1 FPS with PyTorch. 

Our contributions could be summarized as:
\begin{itemize}
\item \textbf{Analysis}: We provide a detailed examination of key components within sparse camera-LiDAR fusion frameworks, identifying limitations through comprehensive experiments and visualizations.
\item \textbf{Improvement}: We introduce specific, performance-optimized solutions that address the identified issues, enhancing the performance and efficiency.
\item \textbf{Performance}: The proposed method, \textbf{FlatFusion}, surpasses existing sparse fusion methods by a notable margin in terms of both accuracy and speed, offering a new standard for the field.
\end{itemize}

\emph{We hope our findings could more provide design insights for sparse camera-LiDAR fusion.}

\addtolength{\textheight}{0cm}   




\section{Related Work}

\subsection{LiDAR-based 3D Perception}
Owing to its capacity for delivering accurate locational data, LiDAR has ascended to a pivotal role as a crucial sensor in ensuring the reliability of autonomous driving systems~\cite{wu2022trajectory,jia2023thinktwice,jia2023driveadapter}. 
Recent LiDAR-centric methods can be broadly categorized into two types according to the form of encoding, i.e., convolution based and Transformer based.
More precisely, earlier methods~\cite{qi2017pointnet,zhou2018voxelnet,lang2019pointpillars} processed point cloud data directly, or transformed it into fixed-size voxels or pillars, then employed 3D CNNs for feature extraction.
Additionally, some methods~\cite{yan2018second,choy20194d,yin2021center,zhang2024hednet,zhang2024safdnet} have utilized sparse convolution techniques to further enhance computational efficiency.
On the other hand, Transformer has recently become popular in the point-cloud field.
PCT~\cite{guo2021pct} first computes self-attention globally over the entire point cloud, but its computational complexity grows quadratically with the number of points, making it difficult to adapt to large-scale point clouds.
SST~\cite{fan2022embracing}, DSVT~\cite{wang2023dsvt} and FlatFormer~\cite{liu2023flatformer} partition the BEV space into non-overlapping and equally sized windows. These windows are then subjected to a shift similar to that of the SwinTransformer~\cite{liu2021swin} to calculate and exchange window self-attention. 
Subsequently, PTv3~\cite{wu2024point} expands the serialized neighbor grouping across multiple tasks.

\subsection{Visual-based 3D Perception}
 Recently, camera-based 3D perception has gained significant interest due to the cost-effectiveness of cameras~\cite{wu2023PPGeo,lu2024activead,jia2020sentimem}.
 LSS~\cite{philion2020lift} spearheaded the process of transforming image features into 3D space by predicting depth maps from images.
 Following this, BEVDet~\cite{huang2021bevdet} and BEVDepth~\cite{li2023bevdepth} advance this concept, further enhancing the performance of camera-only BEV perception. 
 PETR~\cite{liu2022petr} achieves encoding and decoding by integrating 3D spatial positional embeddings into the image features, leveraging the transformer directly.
 Unlike these 2D to 3D transformations, BEVFormer~\cite{li2022bevformer} utilizes deformable attention~\cite{zhu2020deformable} to project BEV features onto the image and computes cross-attention with image features.
 Recently, several approaches~\cite{wu2023heightformer,li2023fb,li2024dualbev} have attempted to combine these two view transformations to achieve more robust BEV representation.
 Furthermore, diverse temporal fusion strategies~\cite{li2022bevformer,huang2022bevdet4d,wang2023exploring} have been extensively employed in camera-only BEV perception.
 Nevertheless, issues such as ambiguity and occlusion of depth remain challenges for pure camera solutions.

\subsection{Multi-modal 3D Perception}
Combining LiDAR's accurate position information and RGB camera's rich semantic information undoubtedly can improve notably for accurate 3D perception and reliable autonomous driving system~\cite{li2024think2drive,jia2025bench2drive,you2024bench2drive}.
Initial multi-modal approaches~\cite{huang2020epnet,vora2020pointpainting,wang2021pointaugmenting} were mainly point-based, i.e., using semantic information from images to enhance the point cloud with richer features.
BEV-based fusion methods~\cite{liu2023bevfusion,liang2022bevfusion} first encode BEV features of each modality and then concatenate them.
In order to prevent the loss of height information, UVTR~\cite{li2022unifying} employs voxel-level modeling and leverages the transformer decoder to sample features in a uniform space.
GAFusion\cite{li2024gafusion} leverages LiDAR guidance to compensate for depth distribution of the camera features. IS-Fusion \cite{yin2024fusion} jointly captures the instance- and scene-level contextual information.
Besides BEV-based fusion, there exist other fusion approaches. For instance, Transfusion~\cite{bai2022transfusion} utilizes image-guided object queries to obtain detection boxes.
DeepInteraction~\cite{yang2022deepinteraction} proposes a bimodal interaction strategy by alternating modalities of cross attention.

As for recent sparse fusion methods: CMT~\cite{yan2023cross}, SparseFusion~\cite{xie2023sparsefusion}, and  ObjectFusion~\cite{cai2023objectfusion} align object-centered features across modalities.
However, they heavily rely on the feature quality of single branch and thus require large backbones (e.g., VoVNet, Swin-Tiny) which is less efficient. UniTR~\cite{wang2023unitr} uniformly adopts windowed attention to fuse information in different representation spaces, but its projection process is rather inaccurate and thus degenerates performance. 

\section{Method}

    \begin{figure}[!t]
        \centering
        \begin{subfigure}[b]{0.48\columnwidth}
            \centering
            \includegraphics[width=\columnwidth]{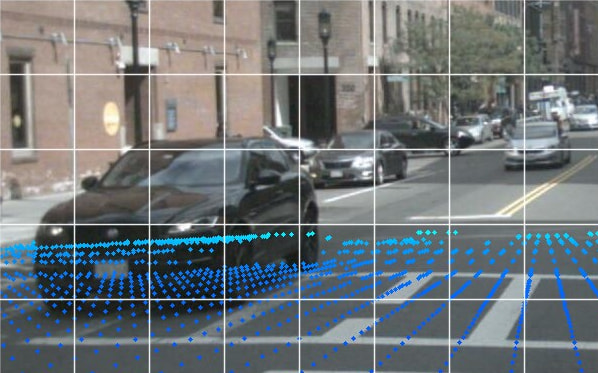}
            \caption{Pillar LiDAR Projection~\cite{wang2023unitr}}
            \label{fig:3d-2d-pillar}
        \end{subfigure}
        \begin{subfigure}[b]{0.48\columnwidth}
            \centering
            \includegraphics[width=\columnwidth]{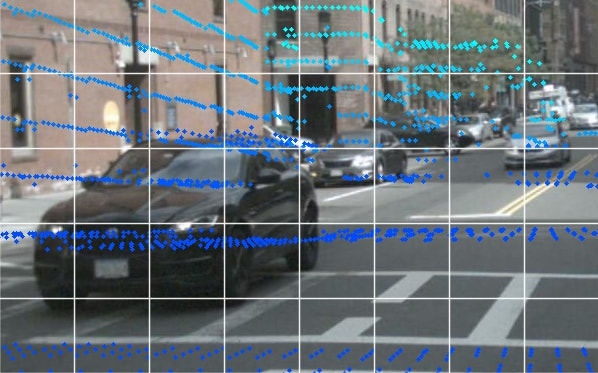}
            \caption{Voxel LiDAR Projection}
            \label{fig:3d-2d-voxel}
        \end{subfigure}
        \caption{\textbf{Comparison of different 3D-to-2D fusion strategies.} Pillar LiDAR feature lose $z$-axis information and thus is only projected to limited area in pixel space, compared to the voxel one. The poor correspondence results in degenerated fusion performance.}
        \label{fig:height-comparison}
        \vspace{-2mm}
    \end{figure}

In this section, we first introduce the general Transformer-based sparse camera-LiDAR fusion framework to study and then we give in-depth investigations about each component.

\subsection{General Framework \& Experiments Setting}

As shown in Fig.~\ref{fig:framework}, sparse Transformer-based camera-LiDAR fusion framework consists of \textit{backbones} for each modality which tokenize the inputs, two representation spaces to fuse information (2D image plane \& 3D space) which requires different \textit{projection strategies}, and sparse windowed attention blocks which require different \textit{neighbor partition strategies} and customized \textit{Transformer structure}.

In the following sections, all experiments are conducted on nuScenes and results on validation set is reported. Latency is tested in RTX4090 with batch size 1 and we report the latency of the compared component only. 
Please refer to Sec.~\ref{sec:exp} for details. 
We ensure that all experiments are strictly controlled, allowing only one variable. Final selection and best results in each column are \textbf{marked}.
    
\subsection{Backbones}

Raw sensor data, i.e., images and point clouds, must be transformed into tokens before being fed into the Transformer. Given the presence of fusion blocks, \textbf{it may not be necessary to use overly complex backbones to achieve the optimal balance between performance and efficiency}. For image backbones, as shown in Table~\ref{tab:backbone} upper part, we compare: (I) 4 Transformer layers (as in UniTR~\cite{wang2023unitr}), (II) ResNet18 (a widely used lightweight encoder), and (III) VoVNet~\cite{lee2019energy} (required by instance query-based methods like CMT~\cite{yan2023cross} and SparseFusion~\cite{xie2023sparsefusion}). All image backbones are ImageNet pretrained while 4 Transformer layers are further pretrained in nuImage by~\cite{wang2023unitr}. For LiDAR backbones, as shown in Table~\ref{tab:backbone} lower part, we compare: (I) VoxelNet~\cite{zhou2018voxelnet} (sparse-conv network), (II) 4 sparse point cloud Transformer layers~\cite{wang2023dsvt,liu2023flatformer}, and (III) None (compensated with 4 additional fusion layers). We conclude:

\begin{itemize}
\item The shallow ViT in~\cite{wang2023unitr} performs poorly due to a mismatch between the lightweight network requirements and the inherently heavy nature of ViTs, which necessitate large models and extensive data.  Compared to ResNet18, VoVNet introduces significant latency and thus we choose ResNet18, which achieves good trade-off.
\item The sparse point cloud Transformer outperforms sparse-conv based VoxelNet. Omitting the LiDAR backbone leads to a substantial performance drop, underscoring the importance of tokenization.
\end{itemize}
    \begin{table}[t]
        \centering
        \caption{Design of single modal tokenizer} 
        \begin{tabular}{llccc}
        \hline
        \toprule
        Modality                  & Type                 & NDS$\uparrow$  & mAP$\uparrow$ & Latency$\downarrow$ \\ \midrule
        \multirow{3}{*}{Image} & Transformer layers~\cite{wang2023unitr}         &   72.3  & 68.5     & 7.9ms        \\
                               & \textbf{ResNet18 + FPN}             &  \textbf{73.4}   & \textbf{70.2}    &  \textbf{6.1ms }      \\ 
                               & VoVNet + FPN             & 73.1    &  70.1  &  71.2ms       \\ \midrule
        \multirow{3}{*}{LiDAR} & VoxelNet           &   73.2    &  \textbf{70.2}   &  16.9ms       \\ 
                               & \textbf{Transformer layers} &  \textbf{73.4}   &   \textbf{70.2}  &  \textbf{10.8ms}       \\
                               & Fusion layers &  72.5   &   69.1  & 17.1ms        \\ \bottomrule
                               \hline
        \end{tabular}
    
    \vspace{-2mm}
    \label{tab:backbone}
    \end{table}
    
\subsection{Fusion \& Projection}

Effective fusion of 2D image features and 3D LiDAR features is central to our framework, yet it is challenged by the disparity between pixel space and voxel space. Thus, it is common practice~\cite{liu2023bevfusion,yang2022deepinteraction} to either project LiDAR features to the image space (3D-to-2D) or project image feature to 3D space (2D-to-3D). This is followed by local neighborhood fusion, as global attention is computationally prohibitive and lacks locality related inductive bias. In the following part, we investigate projecting strategies, the order and necessity of these two types of projections.

\noindent\textbf{3D-to-2D}:
While projecting LiDAR features into pixel space is conceptually straightforward, existing methods such as UniTR~\cite{wang2023unitr} encounter limitations. \textbf{They operate on LiDAR features in the pillar space, which lacks the vertical (z-axis) dimension.} Consequently, as depicted in Fig.~\ref{fig:3d-2d-pillar}, the resulting LiDAR features' neighbors of image features are confined to the ground plane, leading to a significant loss of contextual information. To address this, we propose to utilize voxel-based LiDAR features for 3D-to-2D fusion, as illustrated in Fig.~\ref{fig:3d-2d-voxel}. \textbf{Voxel representations ensures that LiDAR features are comprehensively aligned with the image}, providing a fuller representation of the scene. Towards a fair comparison as the voxel representation introduces extra computations, we also experimented with adding 4 additional layers for the pillar baselines, with results summarized in Table~\ref{tab:projection}. Our findings indicate that voxel-based LiDAR features for 3D-to-2D fusion yield substantial performance improvements over the pillar-based representation, despite a moderate increase in computational demand. Notably, augmenting pillar representation with extra layers did not yield performance gains, reinforcing the efficacy of our voxel-based approach.

    \begin{figure}[!t]
     \centering
     \begin{minipage}[t]{1.0\columnwidth}
       \centering
       \includegraphics[width=\columnwidth]{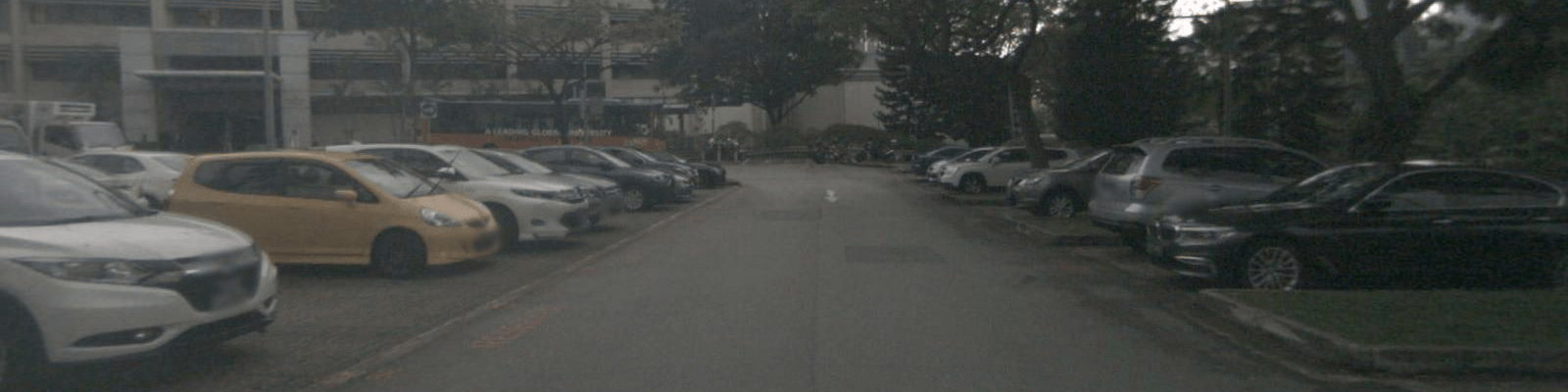}
     \end{minipage}

     \begin{minipage}[t]{1.0\columnwidth}
       \centering
       \includegraphics[width=\columnwidth]{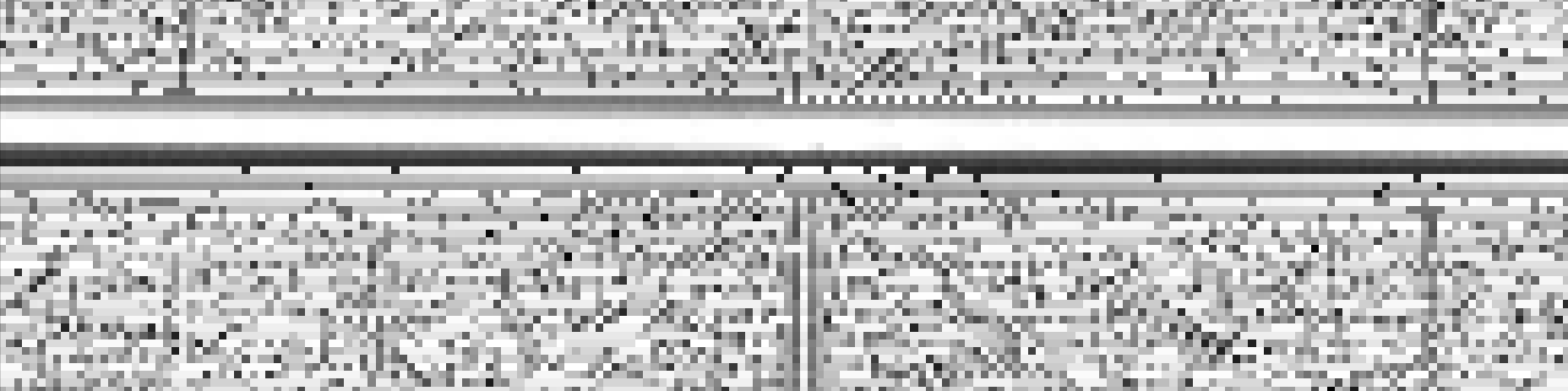}
     \end{minipage}

     \begin{minipage}[t]{1.0\columnwidth}
       \centering
       \includegraphics[width=\columnwidth]{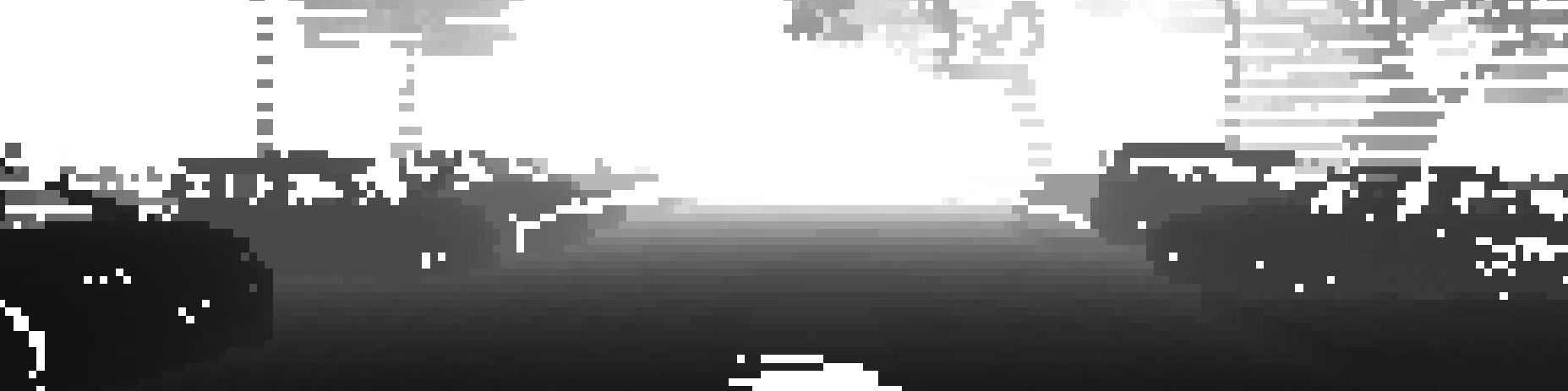}
     \end{minipage}
     \caption{\textbf{Depth map for 2D-to-3D projection with different strategies.} Up to down: Raw image, Nearest Neighbor, Partial Projection. The projected pixel with larger depth would have a lower intensity in visualization. Nearest Neighbor projection has much worse accuracy of depth compared to Partial Projection. The dropped pixels in Partial Projection are mostly background and thus the information lost is limited.\label{fig:2d-to-3d}}
    \end{figure}

    \begin{table}[t]
        \centering
        \caption{Design of projection} 
    \begin{tabular}{llccc}
    \hline
    \toprule
     & Type                 & NDS$\uparrow$  & mAP$\uparrow$ & Latency$\downarrow$ \\ \midrule
    \multirow{3}{*}{3D-to-2D}    &    Pillar        &  72.8   & 69.5    &\textbf{13.2ms}        \\
        &                    \textbf{Voxel}            &  \textbf{73.4}   &  \textbf{70.2}   & 16.1ms        \\ 
        &                   Pillar (2$\times$ layers)             &  72.5   &   69.4  & 18.7ms        \\
    \midrule
    \multirow{3}{*}{2D-to-3D} &    Nearest Neighbor         &   72.9  &  69.8   &  \textbf{7.8ms}     \\
        & Depth Estimation             &  \textbf{73.4}   & 70.0    &  16.1ms       \\ 
       & \textbf{Partial Projection}             & \textbf{73.4}    &  \textbf{70.2}   & 10.7ms        \\
    \bottomrule
    \hline
    \end{tabular}
    \vspace{-2mm}
    \label{tab:projection}
    \end{table}

\noindent\textbf{2D-to-3D}: Projecting image features to LiDAR space is complicated by the inherent depth ambiguity in pixels~\cite{park2023depth}. Therefore, a well-designed projection strategy is required. We investigate three approaches: (I) \textbf{Nearest Neighbors}: projecting according to the nearest voxels in pixel space~\cite{yin2021multimodal,wang2023unitr}. While avoiding any learnable modules, this method is highly inaccurate because nearby voxels in 2D space may correspond to very different depths, introducing significant errors, as shown in Fig.~\ref{fig:2d-to-3d} (Middle). 
(II) \textbf{Depth Estimation}: training an additional depth network and using the predicted depth for projection. It can be accurate, but requiring burdensome network~\cite{lee2019energy} to achieve accurate depth prediction.
(III) \textbf{Partial Projection}:  Unlike existing designs, \textbf{we propose to only project those image patches that have LiDAR points on, ensuring accurate depth}. Table~\ref{tab:projection} shows that Partial Projection is significantly faster than Depth Estimation, while  Nearest Neighbors degrades performance.  As shown in Fig.~\ref{fig:2d-to-3d} (Lower), since image features are usually downsampled, major part of them are projected and thus the information loss is limited. Besides, all image features are used in 3D-to-2D blocks, which could compensate for loss of some pixels and hence we choose accurate and efficient Partial Projection.


\noindent\textbf{Order \& Necessity}: We explore the sequence in which 2D-to-3D and 3D-to-2D fusion projections are applied: \emph{2D-to-3D first} or \emph{3D-to-2D first}. Additionally, we conduct experiments with single fusion projections while maintaining a consistent number of Transformer layers to isolate the impact. Our results, as in Table~\ref{tab:order-necessity}, underscore the importance of both fusion projections, which both contribute unique perspectives into the multi-modal association, aligned with findings in~\cite{yang2022deepinteraction,wang2023unitr}. Notably, \emph{3D-to-2D first} outperforms \emph{2D-to-3D first} approach by a large margin, which can be attributed to the fact that LiDAR voxel features undergo downsampling between stages~\cite{zhou2018voxelnet,wang2023dsvt}, whereas the z-axis information, crucial for accurate 3D-to-2D fusion, is lost in the downsampling process. \textbf{\emph{3D-to-2D first} provides more fine-grained LiDAR voxel features in z-axis for 3D-to-2D fusion and thus achieves best performance.}

    \begin{table}[t]
        \centering
        \caption{Design of order and necessity} 
    \begin{tabular}{lccc}
    \hline
    \toprule
     Type                 & NDS$\uparrow$  & mAP$\uparrow$ & Latency$\downarrow$ \\ \midrule
    \textbf{3D-to-2D First}         &  \textbf{73.4}   &   \textbf{70.2}  & 27.5ms        \\
        2D-to-3D First            &  72.9   & 69.5    & 29.2ms        \\ 
       3D-to-2D Only          & 71.8    &  67.8   & 25.4ms        \\
       2D-to-3D Only             &   72.5  &  69.1   & \textbf{24.6ms}        \\
    \bottomrule
    \hline
    \end{tabular}
    \vspace{-2mm}
    \label{tab:order-necessity}
    \end{table}

\subsection{Partition}
\begin{figure}[!t]
        \centering
        \captionsetup{skip=5pt}
        \begin{subfigure}[b]{0.32\columnwidth}
            \centering
            \includegraphics[width=1.0\columnwidth]{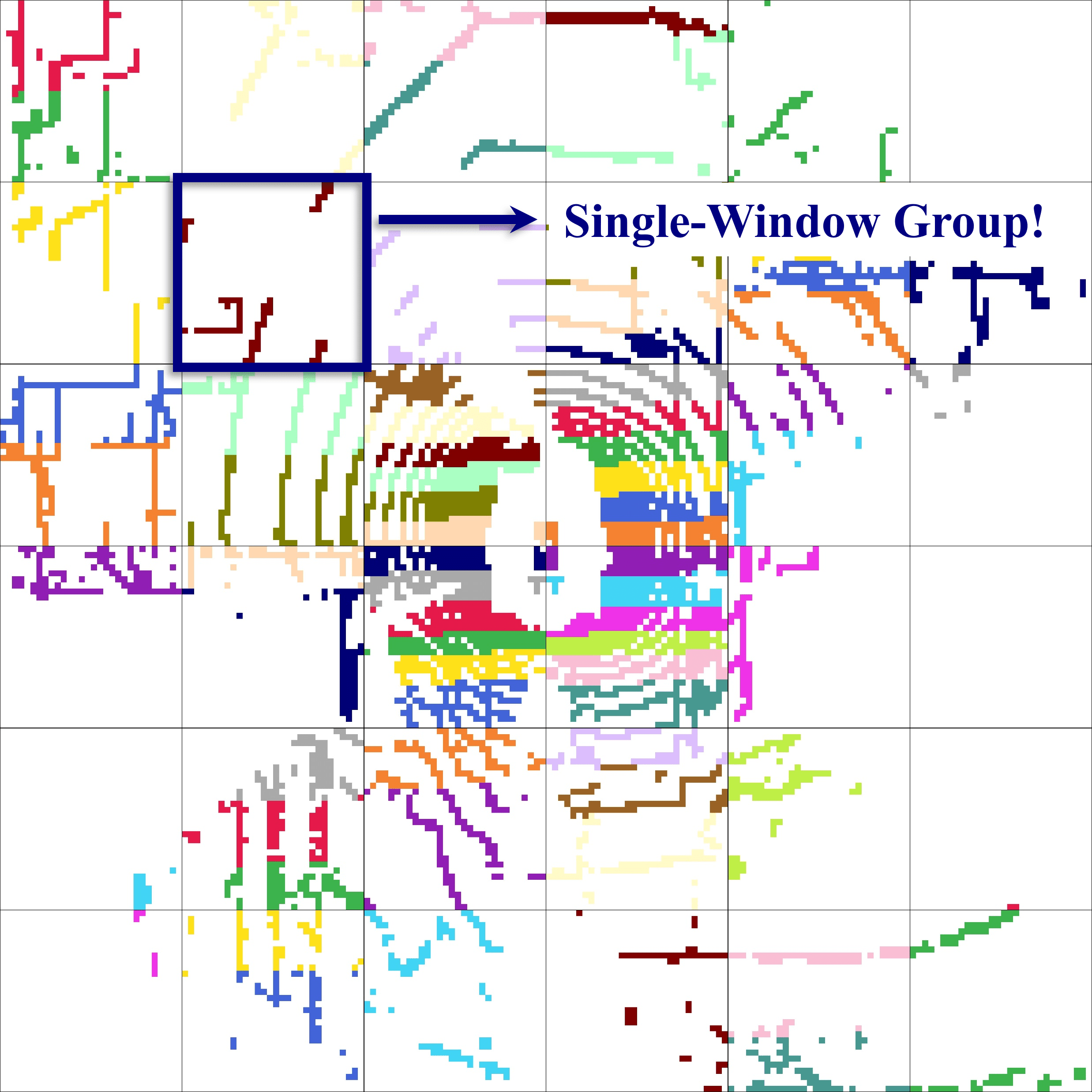}
            \caption{Dynamic Set} 
        \end{subfigure}
        \begin{subfigure}[b]{0.32\columnwidth}
            \centering\includegraphics[width=1.0\columnwidth]{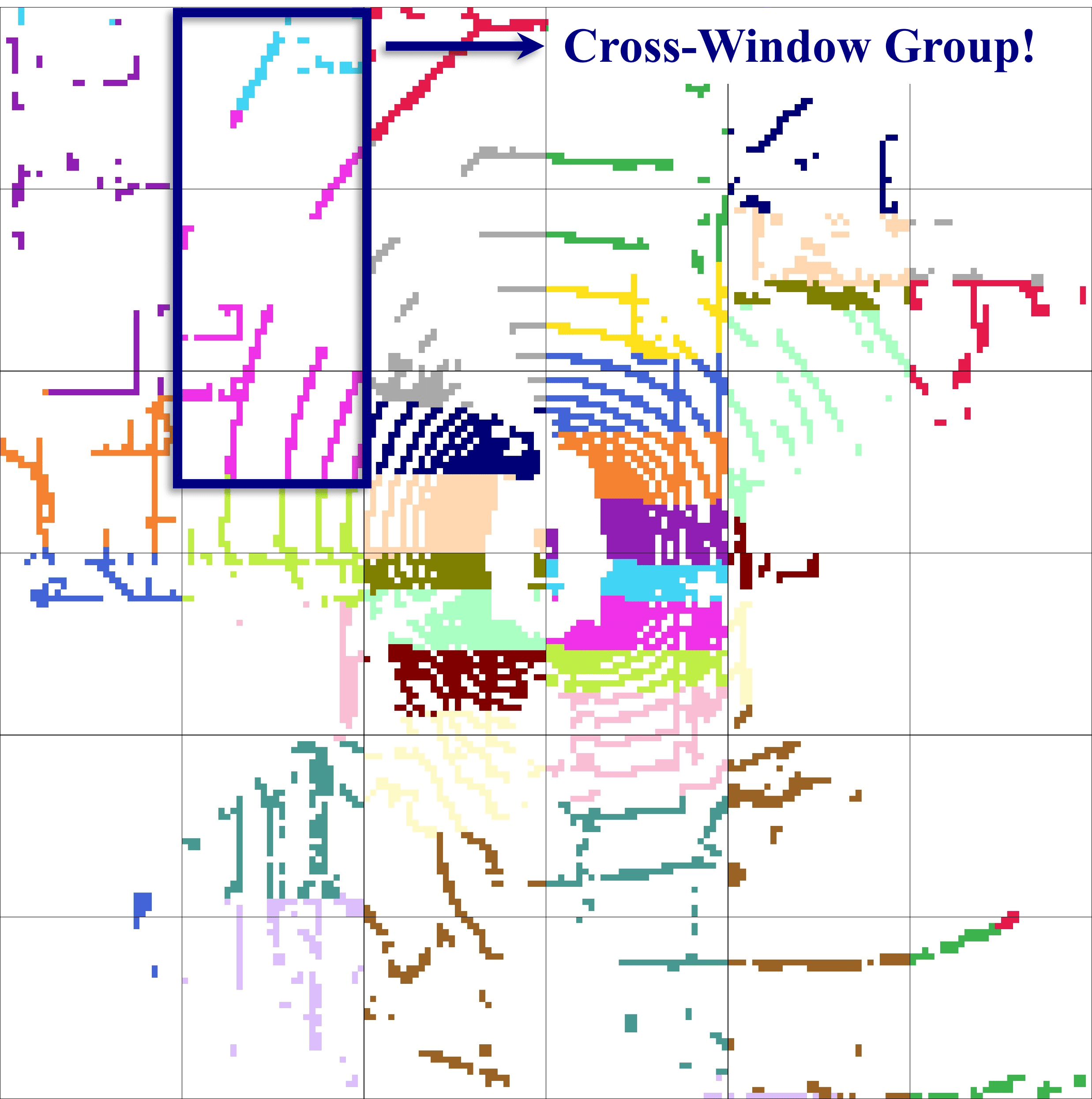}
            \caption{Flatten Window}
        \end{subfigure}
        \begin{subfigure}[b]{0.32\columnwidth}
            \centering
            \includegraphics[width=1.0\columnwidth]{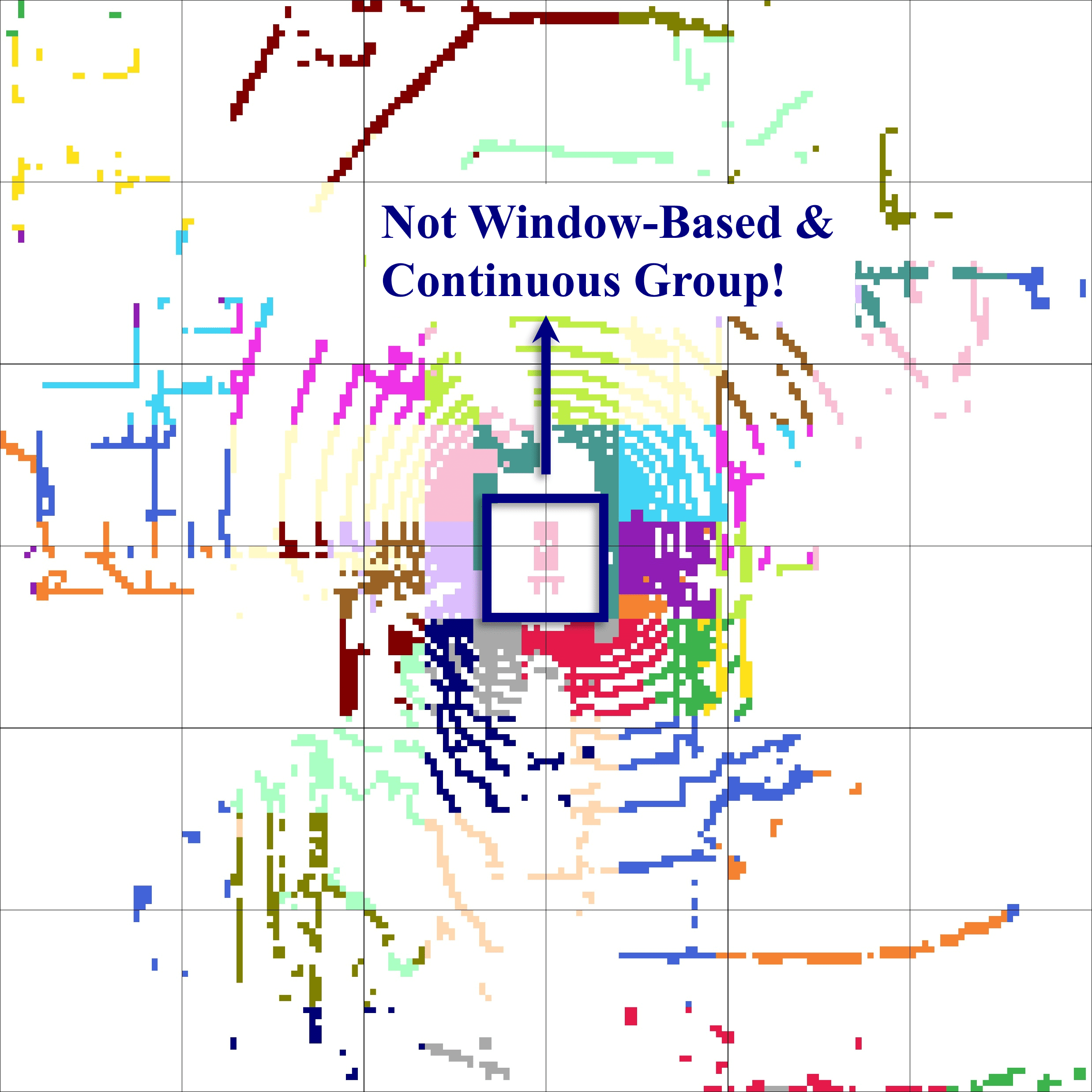}
            \caption{Space-Filling}
        \end{subfigure}

        \begin{subfigure}[b]{0.32\columnwidth}
            \centering
            \includegraphics[width=1.0\columnwidth]{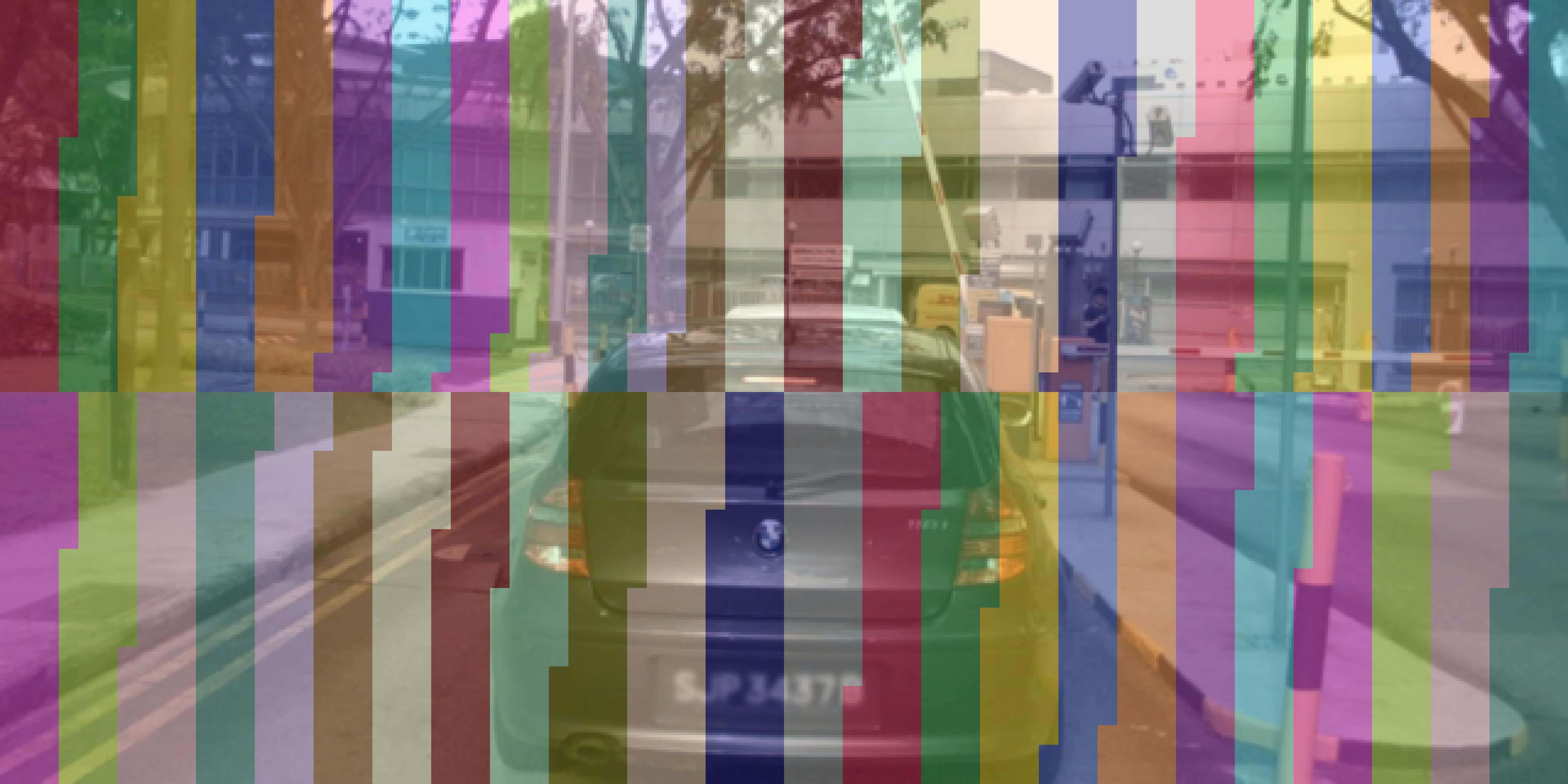}
            \caption{Dynamic Set}
        \end{subfigure}
        \begin{subfigure}[b]{0.32\columnwidth}\centering\includegraphics[width=1.0\columnwidth]{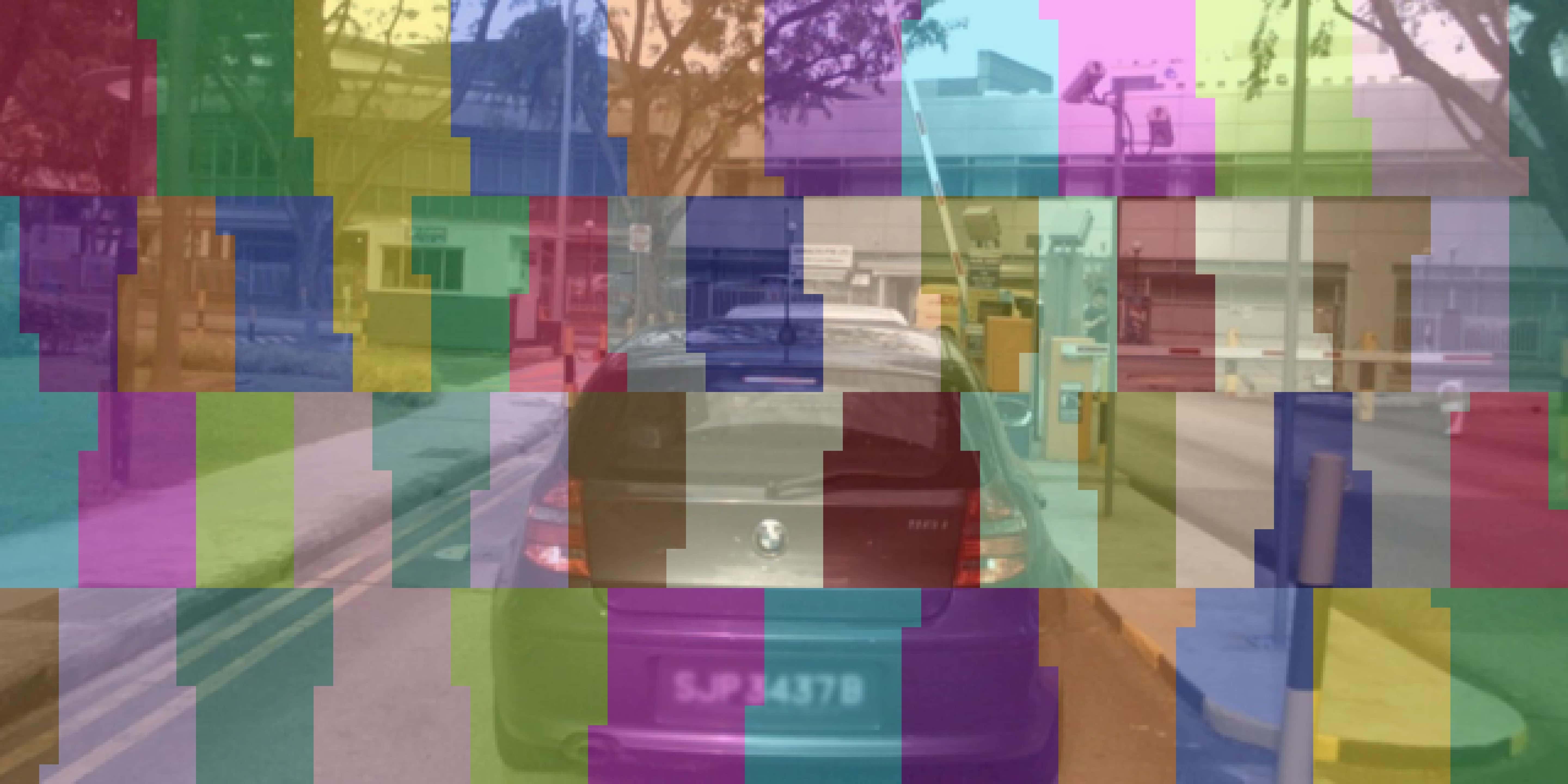}
            \caption{Flatten Window}
        \end{subfigure}
        \begin{subfigure}[b]{0.32\columnwidth}
            \centering
            \includegraphics[width=1.0\columnwidth]{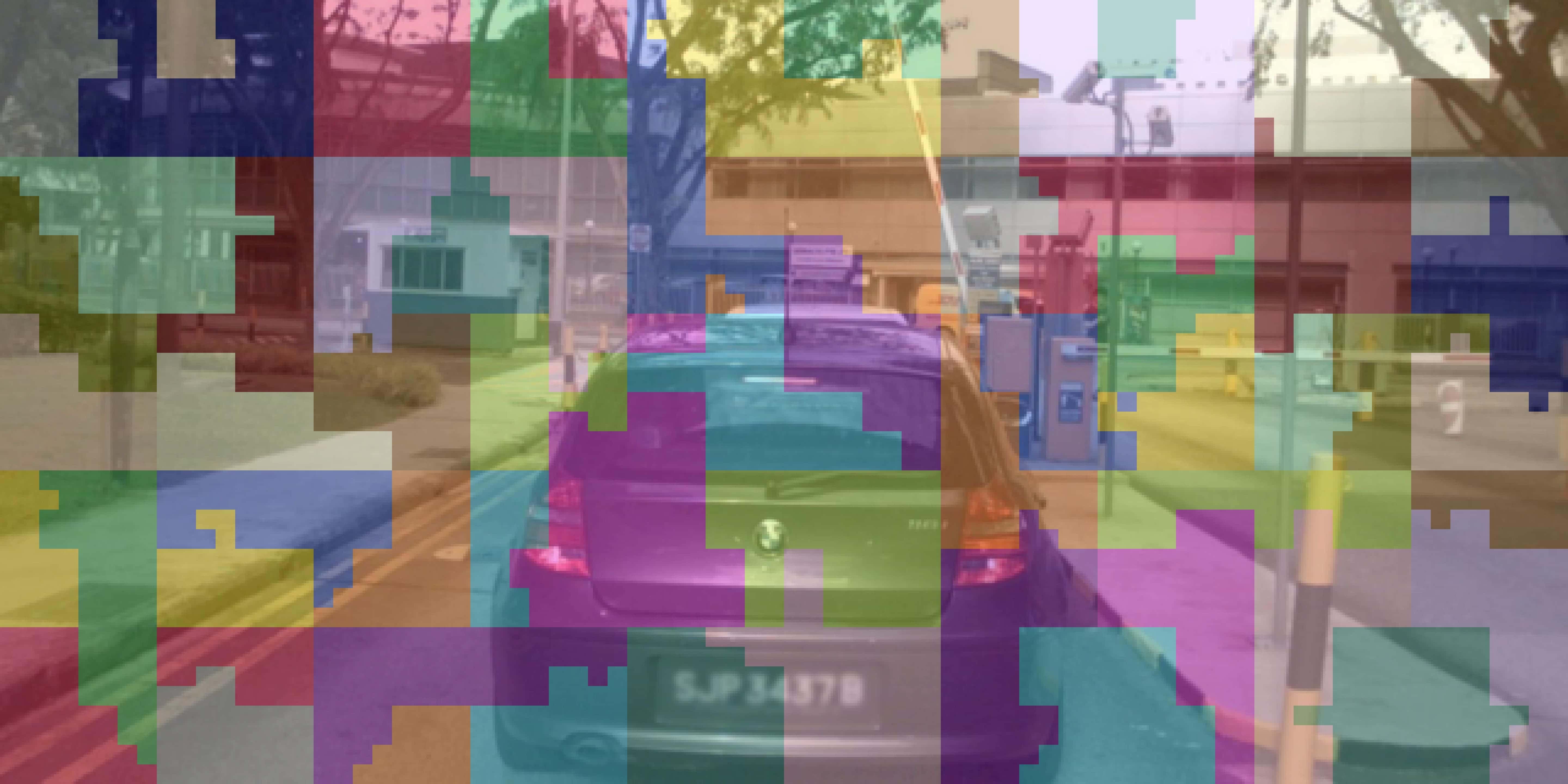}
            \caption{Space-Filling}
        \end{subfigure}
        \caption{\textbf{Comparison of different partition strategies.} Up to down: 3D space, perspective space. For 3D space, each grid represents a window in Dynamic Set and Flatten Window. Tokens of the same color within a localized area represent that they are partitioned into the same group to conduct attention. Dynamic Set ensures all elements of the same group are within a single window and thus preserve strong locality while in Flatten Window, elements could be cross multiple windows. Space-Filling Curve uses mathematical curves for indexing without considering specific windows, which possesses certain locality yet without any hard constraint and thus is efficient. However, its pattern in perspective space is irregular.}
        \label{fig:partition}
    \end{figure}

Unlike images or languages, 3D data, e.g. point clouds, is naturally sparse~\cite{spconv2022}, especially in the autonomous driving scenarios. Consequently, applying attention necessitates sparse and local windowed mechanism. Instead of partitioning elements into windows of equal shapes as in SST~\cite{fan2022embracing}, state-of-the-art sparse point cloud Transformer~\cite{wang2023dsvt,liu2023flatformer,wu2024point} divides elements into groups of equal sizes, resulting in significantly reduced latency. \textbf{When it comes to fusion, the tokens are also distributed irregularly, which aligns well with the sparse point cloud Transformer.} The extra image modality could render the original conclusions in~\cite{liu2023flatformer,wang2023dsvt,wu2024point} inaccurate.
To this end, we compare three state-of-the-art partition algorithms in the context of fusion: (I) \textbf{Dynamic Set}~\cite{wang2023dsvt}: (I) It first divides the space into several windows based on coordinate and then partitions groups based on the number of non-empty voxels within each window. It aligns tensor shapes by padding and performs attention with masking. (II) \textbf{Flatten Window}~\cite{liu2023flatformer}: It also splits windows but partitions groups based solely on indices, making it much faster than Dynamic Set at the expense of lower locality. (III) \textbf{Space-Filling Curve}~\cite{peano1990courbe,wu2024point}: It uses mathematical curves that traverse every point in a higher-dimensional space, preserving spatial proximity better in 3D space than Flatten Window. In Fig~\ref{fig:partition}, we visualize the three partition algorithms in 3D and 2D space respectively. We can observe that for spatial proximity in 3D space: Dynamic Set $>$ Space-Filling Curve $>$ Flatten Window. In 2D space, Space-Filling Curve is a little bit messy possibly because the mathematical curves is designed for high-dimensional space. 

Table~\ref{tab:partition} gives performance comparisons, leading to the following conclusions:
\begin{itemize}
\item Dynamic Set is the slowest due to its complex partitioning, even with customized CUDA operators. Flatten Window is the fastest as it relies on simple indexing. Besides, it does not require attention mask and thus supports 
FlashAttention~\cite{dao2022flashattention}, which brings further speedup.
\item Dynamic Set and Flatten Window has similar performance, indicating the robustness of attention mechanism while Space-Filling Curve performs worst due to their bad 2D partitioning.

\end{itemize}

\begin{figure}[t]
        \centering
        \captionsetup{skip=5pt}
        \begin{subfigure}[b]{0.32\columnwidth}
            \centering
            \includegraphics[width=1.0\columnwidth]{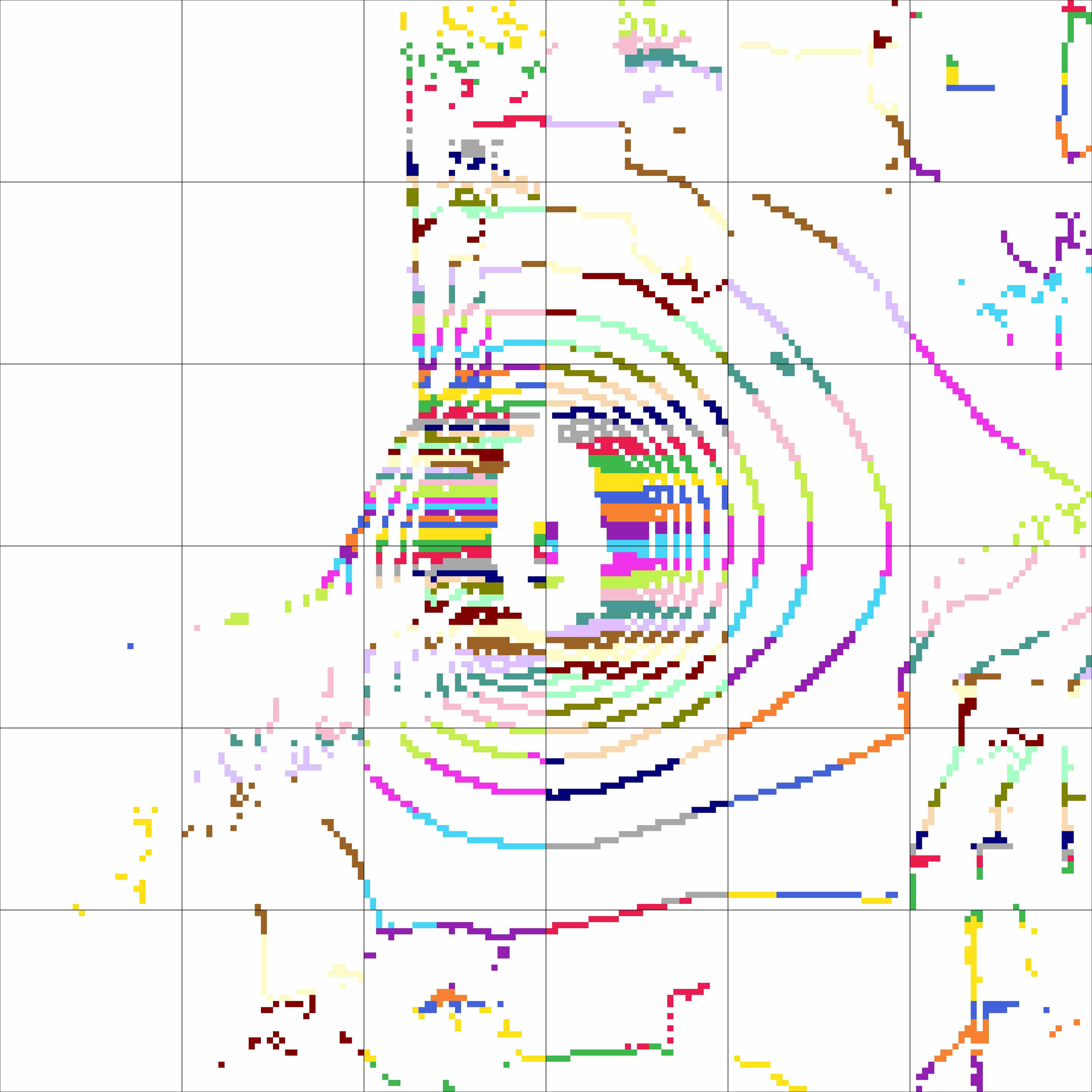}
            \caption{Group size: 40}
        \end{subfigure}
        \begin{subfigure}[b]{0.32\columnwidth}
            \centering\includegraphics[width=1.0\columnwidth]{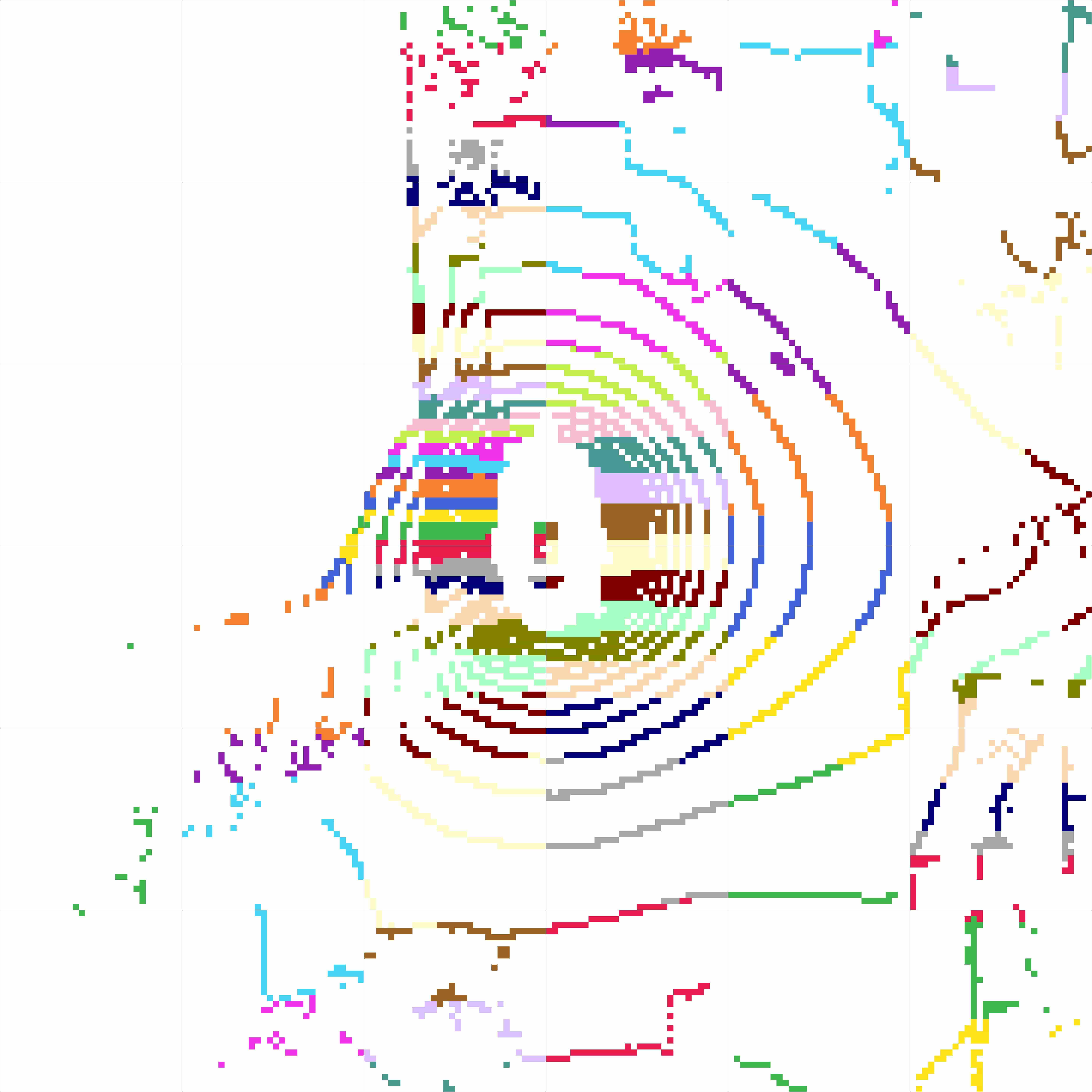}
            \caption{Group size: 80}
        \end{subfigure}
        \begin{subfigure}[b]{0.32\columnwidth}
            \centering
            \includegraphics[width=1.0\columnwidth]{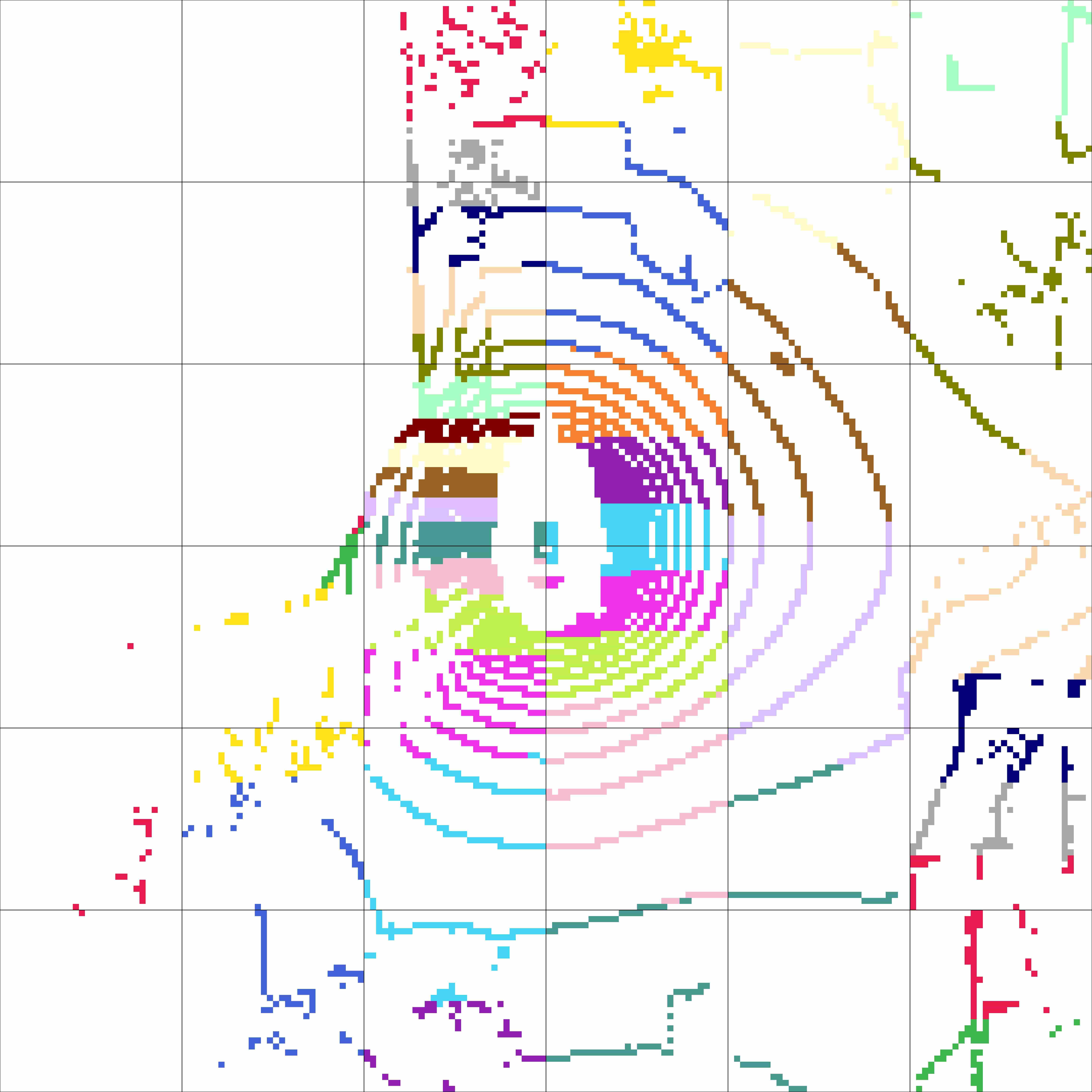}
            \caption{Group size: 160}
        \end{subfigure}
        \caption{Visualization of different group size in 3D space.\label{fig:group}}
    \end{figure}

\noindent\textbf{Group Size}: One key hyper-parameter of sparse windowed attention is the group size, i.e., the number of tokens per set to conduct attention. We investigate the performance of group size under 3 scales: 40, 80, 160, as in Table~\ref{tab:partition}. We find that 80 is suitable, which is quite different from the conclusion in PTv3~\cite{wu2024point}, where they find larger group size up to 1024 could consistently bring performance gains. The discrepancy could potentially be attributed to their primary focus on indoor point clouds~\cite{dai2017scannet}, which are typically denser than their outdoor counterparts~\cite{caesar2020nuscenes}. We visualize the partitioning results of different group size in Fig.~\ref{fig:group}. We can find that \textbf{when using small group size, the receptive field is limited while large group would leads to bad locality}, indicating the necessity of tuning.


\subsection{Transformer Structure}
Transformer~\cite{vaswani2017attention} consists of attention, FFN, residual link, LayerNorm, and position encoding (PE). Following LLaMA family~\cite{touvron2023llama}, we adopt FlashAttention~\cite{dao2022flashattention} and SwiGLU~\cite{shazeer2020glu}. Existing works like FlatFormer~\cite{liu2023flatformer} and UniTR~\cite{wang2023unitr} both adopt \emph{PostNorm}~\cite{vaswani2017attention} while modern architecture~\cite{touvron2023llama} usually adopts \emph{PreNorm}~\cite{chen2018best}. UniTR proposes to use an extra \emph{block-wise residual} and post-norm. As for PE, we examine window-level local indices~\cite{wang2023unitr} or 3D position encoding. In Table~\ref{tab:structure}, we find that PreNorm \& 3D PE achieves best performance, aligned with modern findings~\cite{touvron2023llama}.

    \begin{table}[t]
        \centering
        \caption{Design of partition} 
    \begin{tabular}{lcccc}
    \hline
    \toprule
     & Type                 & NDS$\uparrow$  & mAP$\uparrow$ & Latency$\downarrow$ \\ \midrule
    \multirow{3}{*}{Algorithm} & Dynamic Set         &   \textbf{73.4}  &  70.1   &  28.5ms       \\
    & \textbf{Flatten Window}  &  \textbf{73.4}   & \textbf{70.2}     & \textbf{17.3ms}         \\ 
    & Space-Fill Curves      & 73.1    &  70.0   & 22.6ms        \\
    \midrule
    \multirow{3}{*}{Group Size} &    40         &  72.7   &  69.7   &  25.3ms       \\
    & \textbf{80}      &   \textbf{73.4}  &   \textbf{70.2}  & \textbf{22.7ms}        \\ 
    & 160          &  72.8   & 69.6    &  23.9ms      \\ 
    \bottomrule
    \hline
    \end{tabular}
    \label{tab:partition}
    \end{table}

   \begin{table}[!t]
        \centering
        \caption{Design of transformer structure\label{tab:structure}} 
    \begin{tabular}{lccc}
    \hline
    \toprule
     Type                 & NDS$\uparrow$  & mAP$\uparrow$ & Latency$\downarrow$ \\ \midrule
    PostNorm             &   73.0  & 69.8   &    \textbf{48.1ms}     \\ 
    Block Residual~\cite{wang2023unitr}            &   72.6  &   69.3  &   49.2ms      \\
    Window-Level PE~\cite{wang2023unitr}             &   72.5  & 69.3    &     52.0ms 
    \\
    \textbf{PreNorm + 3D PE}        & \textbf{73.4}     &   \textbf{70.2}  &    48.8ms     \\ 
    \bottomrule
    \hline
    \end{tabular}
    \vspace{-2mm}
    \end{table}

\begin{table*}[!t]
    \centering
    \caption{\textbf{Performance of 3D detection on nuScenes validation and test set.} FPS is measured on A100 GPU by default with cached intrinsic and extrinsic.  $^\dag$: FPS on RTX3090. $^\ddag$: FPS on RTX4090. \label{tab:val}}
    \begin{tabular}{l|c|c|c|cc|cc|c}
    \hline
    \toprule
    Method  & Present at  & Camera Branch & Fusion Strategy  & NDS (\textit{val}) & mAP (\textit{val}) & NDS (\textit{test}) & mAP (\textit{test}) & FPS \\
    \midrule
    TransFusion~\cite{bai2022transfusion}          & CVPR’22    & ResNet-50~\cite{He_2016_CVPR} & Instance Query & 71.3 & 67.5  &  71.6  &    68.9  & 6.1    \\
    UVTR~\cite{li2022unifying}                     & NeurIPS’22 & ResNet-101~\cite{He_2016_CVPR}  & BEV Concat & 70.2 & 65.4 & 71.1  &  67.1   &  1.8$^\dag$   \\
    BEVFusion~\cite{liang2022bevfusion}            & NeurIPS’22 & Dual-Swin-T~\cite{liang2022cbnet} & BEV Concat  & 71.0    & 67.9 &  71.8 &  69.2  & 0.8    \\
    DeepInteraction~\cite{yang2022deepinteraction} & NeurIPS’22 & ResNet-50~\cite{He_2016_CVPR}  & Dual Attention & 72.6 & 69.9  & 73.4  &  70.8  & 1.8    \\
    FUTR3D~\cite{chen2023futr3d}                   & CVPR’23    & ResNet-101~\cite{He_2016_CVPR} & Instance Query & 68.3  & 64.5  &  - &  -  &  3.3   \\
    MSMDFusion~\cite{jiao2023msmdfusion}           & CVPR’23    & ResNet-50~\cite{He_2016_CVPR}  & BEV Concat & 72.1  & 69.3  & 74.0   &  71.5  & 2.1$^\dag$   \\
    BEVFusion~\cite{liu2023bevfusion}              & ICRA’23    & Swin-T~\cite{liu2021swin} & BEV Concat & 71.4  & 68.5  &  72.9 &   70.2  &  7.7  \\
    ObjectFusion~\cite{cai2023objectfusion}        & ICCV’23    & Swin-T~\cite{liu2021swin} & Instance Query & 72.3  & 69.8 &  73.3  &  71.0   & 3.6\\
    SparseFusion~\cite{xie2023sparsefusion}        & ICCV’23    & ResNet-50~\cite{He_2016_CVPR} & Instance Query & 72.8   & 70.4   &  73.8  &   \textbf{72.0}  &  5.6   \\
    CMT~\cite{yan2023cross}                        & ICCV’23    & VoVNet-99~\cite{lee2019energy} & Instance Query & 72.9    & 70.3   &  74.1 & \textbf{72.0}   & 6.4    \\
    FocalFormer3D ~\cite{chen2023focalformer3d}    & ICCV’23    & ResNet-50~\cite{He_2016_CVPR}  & BEV Concat & 73.1           & 70.5   & 73.9   &  71.6   & -    \\
    UniTR ~\cite{wang2023unitr}                    & ICCV’23    & DSVT~\cite{wang2023dsvt} & Dual Attention & 73.3  & 70.5   & 74.5   &  70.9   & 9.3 (8.7$^\ddag$)    \\ 
    FSF~\cite{li2024fully}  & TPAMI’24   & ResNet-50~\cite{He_2016_CVPR} & Instance Query & 72.7  & 70.4   &  74.0 &   70.6 &  7.0$^\dag$   \\ 
    GraphBEV~\cite{song2024graphbev}               & ECCV’24    & Swin-T~\cite{liu2021swin} & BEV Concat & 72.9          & 70.1   &  73.6 & 71.7         & 7.1    \\
    \textbf{FlatFusion}                            & -          & ResNet-18~\cite{He_2016_CVPR}  & Dual Attention & \textbf{73.7} & \textbf{70.6}& \textbf{74.7}  &  71.1  &  \textbf{10.1}$^\ddag$   \\ 
    \bottomrule
    \hline
    \end{tabular}
\end{table*}

\subsection{Final Model} 
The final model could be summarized as: given images from surrounding cameras and point cloud from LiDAR as inputs, we use \textbf{ResNet18}~\cite{He_2016_CVPR} followed by FPN~\cite{lin2017feature} and Dynamic VFE~\cite{zhou2020end} followed by \textbf{Flatten Window} Transformer layers~\cite{wu2024point} as tokenizer to obtain camera tokens with a shape of $N_C*C$ and LiDAR tokens with a shape of $N_L*C$ respectively. Then, we project LiDAR tokens to the pixel space and feed the mixed image and LiDAR tokens into 4 \textbf{Flatten Window} Transformer layers for 3D-to-2D fusion. After that, we project image tokens to 3D space by \textbf{Partial Projection} and use 4 \textbf{Flatten Window} Transformer layers for 2D-to-3D fusion. Finally, we densify~\cite{liu2023flatformer,wang2023unitr} the LiDAR feature followed by several convs and feed it into the 3D detection head~\cite{bai2022transfusion}. Similar to~\cite{wang2023unitr}, optionally, the image features could be  projected by LSS~\cite{philion2020lift,huang2023detecting} to BEV and concatenated with the densified LiDAR features, which leads to better performance in the cost of more latency. The LiDAR feature is downsampled in $z$-axis with attentive pooling~\cite{wang2023dsvt} between stages. All Transformer layers are \textbf{PreNorm} and use \textbf{3D PE} with a \textbf{group size 80}. 

\section{Experiments}
\label{sec:exp}
\subsection{Setup}
\noindent\textbf{Dataset}:
We leverage nuScenes~\cite{caesar2020nuscenes}, a comprehensive public dataset meticulously curated for autonomous driving applications, to assess our performance.
It includes 1,000 scenes, each 20 seconds long, with six cameras and one LiDAR sensor.  We use the official train-val-test split (750-150-150).
This multi-sensor setup captures a wide array of annotations, including 23 object classes and 8 attributes, over 1.4 million annotated 3D bounding boxes.

\noindent\textbf{Metric}: For 3D object detection, we employed two primary metrics: Mean Average Precision (mAP) and the nuScenes Detection Score (NDS). 
mAP is calculated by averaging over the distance thresholds of 0.5$m$, 1$m$, 2$m$, and 4$m$ across all categories.
NDS, the official metric, is a weighted average of mAP and five metrics that measure translation, scaling, orientation,
velocity, and attribute errors.

\noindent\textbf{Implementation}: We use OpenPCDet~\cite{openpcdet2020} as code base and follow the data preprocessing, augmentation, postprocessing, optimizer, and training schedule in UniTR~\cite{wang2023unitr}. In all experiments, the input is the point cloud and all 6 camera images with the size of $256\times704$. The grid size is set to (0.3$m$, 0.3$m$, 0.25$m$) and the hidden dimension is 128. All models are trained with $8\times$ RTX4090.

\begin{figure}[!t]
    \centering
    \includegraphics[width=\columnwidth]{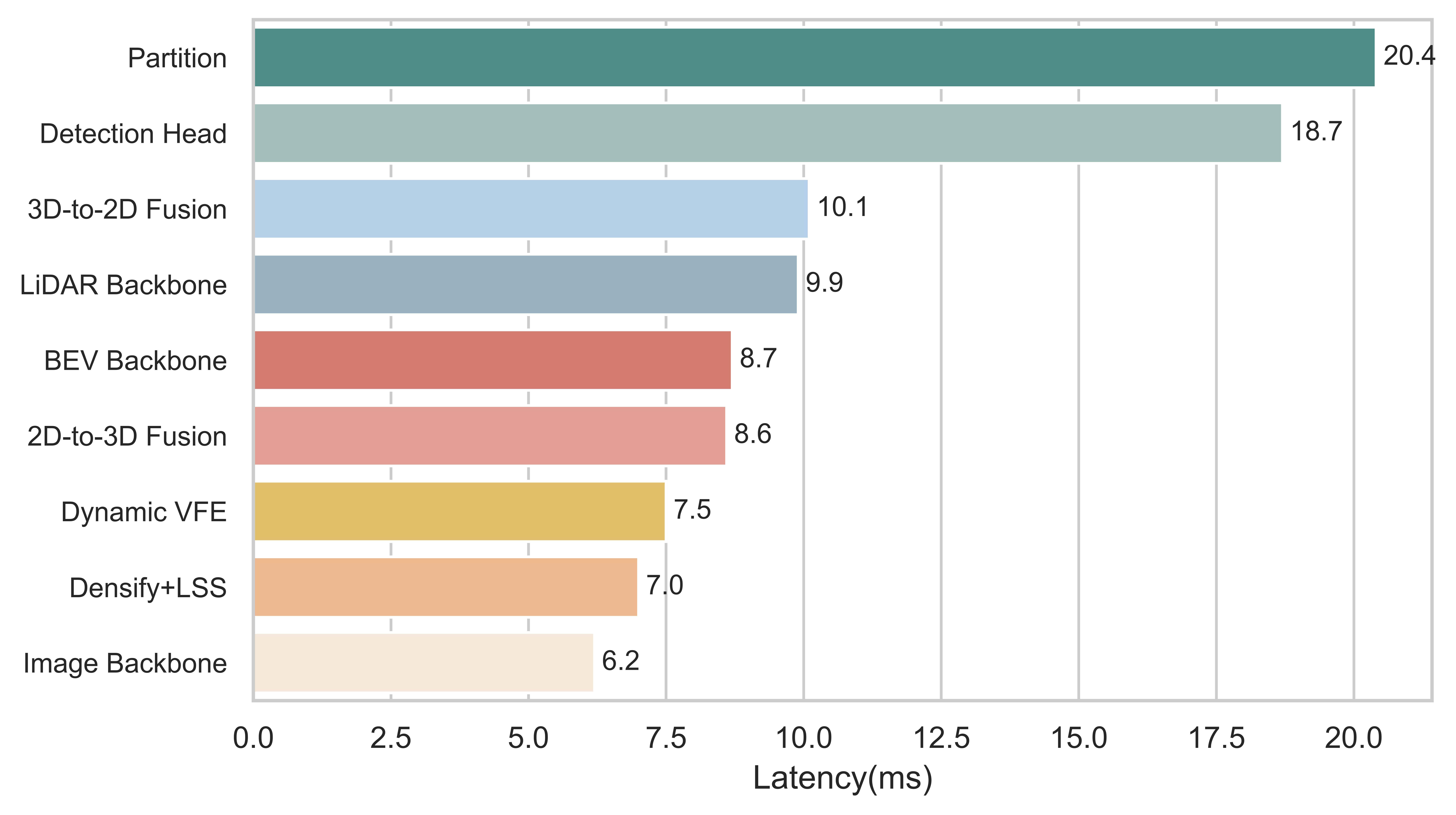}
    \caption{Latency analysis for FlatFusion on the nuScenes \textit{validation} set, measured with an RTX 4090.}
    \label{fig:latency}
\end{figure}

\subsection{Results}

We give results of 3D objection detection on nuScenes validation and test set in Table~\ref{tab:val}.
We could observe that \textbf{FlatFusion} achieves strong performance yet with a high FPS, demonstrating the efficacy of the explored designs. Notably, compared to the latest sparse Transformer based fusion methods, i.e.,  CMT~\cite{yan2023cross}, SparseFusion~\cite{xie2023sparsefusion}, UniTR~\cite{wang2023unitr}, FlatFusion demonstrate explicit performance advantages.

\subsection{Influence of Network Depth}
In Table~\ref{tab:depth}, we study the influence of network depth. As for the number of Transformer layers per block (LiDAR backbone, 3D-to-2D, 2D-to-3D), we find that 2 layers leads to underfit while 8 layers causes unstable training. Regarding BEV encoder, we find that the commonly used SECOND~\cite{yan2018second} performs worse than customized ResNet~\cite{wang2023unitr}, indicating the efficacy of residual connection.

   \begin{table}[!t]
    \centering
    \caption{Influence of network depth\label{tab:depth}} 
    \begin{tabular}{ccccc}
    \hline
    \toprule
      &  Type     &    NDS$\uparrow$  & mAP$\uparrow$ & Latency$\downarrow$     \\ \midrule
    \multirow{3}{*}{Block Depth}  & 2    &  72.4      & 69.0 &  \textbf{39.8 }\\
      & \textbf{4}    &   \textbf{73.4}     & \textbf{70.2} & 48.2  \\
      & 8    &   72.6     &  69.6 &  58.1 \\
    \midrule
    \multirow{3}{*}{BEVEncoder} & SECOND (5+5)~\cite{liu2023bevfusion}   &      72.5  & 69.3 & 6.6  \\
     & SECOND (8+8)~\cite{huang2023detecting}   &  72.8      &  69.3 & 9.2  \\
     & \textbf{ResNet14}~\cite{wang2023unitr}    &  \textbf{73.4 }     & \textbf{70.2} & \textbf{8.7}  \\
     \bottomrule
     \hline
    \end{tabular}
    \end{table}
    
\subsection{Latency Analysis}
In Fig.~\ref{fig:latency}, we give the latency of each component in FlatFusion. We could observe that even though heavily optimized, the partition time is still one of the biggest overhead, which worth further study. Besides, the BEV backbone, as well as the densification and detection head, could potentially be further light-weighted.

\section{Conclusion}
In this work, we delve into the details of sparse Transformer based camera-LiDAR fusion framework. We analyze the issues within existing methods and propose corresponding solutions. The proposed FlatFusion achieves state-of-the-art performance with a high FPS. We aspire for FlatFusion to stimulate future research endeavors centered around the design of efficient and precise multi-sensor fusion transformers.

{
\small
\bibliographystyle{abbrv}
\bibliography{mybibliography}
}

\end{document}